\documentclass{article}

\PassOptionsToPackage{numbers, compress, sort}{natbib}

\usepackage[final]{neurips_2022}

\usepackage{wrapfig}
\usepackage{amsmath,amsfonts,amssymb,amsthm,mathrsfs,bbm,bm,multicol}
\usepackage[british]{babel}
\usepackage{booktabs} 
\usepackage{enumitem}
\usepackage[utf8]{inputenc} 
\usepackage[T1]{fontenc}    
\usepackage{graphicx}
\usepackage[colorlinks = true,
            linkcolor = magenta,
            urlcolor  = blue,
            citecolor = black,
            anchorcolor = blue]{hyperref}
\usepackage{nicefrac}       
\usepackage{microtype}      
\usepackage{physics}
\usepackage{color}
\usepackage{sidecap}
\usepackage{xcolor}

\graphicspath{ {./images/} }

\usepackage[export]{adjustbox}
\usepackage{subfig}
\usepackage{threeparttable}

\usepackage[normalem]{ulem}

\title{Redundant representations help generalization in wide neural networks}

\author{%
   Diego Doimo\thanks{ddoimo@sissa.it} \\
   International School for Advanced Studies \\
   \And
   Aldo Glielmo \\
   International School for Advanced Studies \\
   Bank of Italy\thanks{The views and opinions expressed in this paper are those of the authors and do not necessarily reflect the official policy or position of Bank of Italy.} \\
   \And
   Sebastian Goldt \\
   International School for Advanced Studies \\
   \And
   Alessandro Laio \\
   International School for Advanced Studies
}

\begin{document}
\maketitle

\begin{abstract}
Deep neural networks (DNNs) defy the classical bias-variance trade-off: adding parameters to a DNN that interpolates its training data will typically improve its generalization performance. Explaining the mechanism behind this ``benign overfitting'' in deep networks remains an outstanding challenge. Here, we study the last hidden layer representations of various state-of-the-art convolutional neural networks and find that  if the last hidden representation is wide enough, its neurons tend to split into groups that carry identical information and differ from each other only by statistically independent noise. The number of such groups increases linearly with the width of the layer, but only if the width is above a critical value. We show that redundant neurons appear only when the training is regularized and the training error is zero.
\end{abstract}

\section{Introduction}

Deep neural networks (DNN) have enough parameters to achieve zero
training error, even with random labels~\citep{zhang2016understanding, arpit2017closer}.
In defiance of the classical bias-variance trade-off, the performance of these
\emph{interpolating classifiers} improves as the number of parameters increases well beyond the number of training
samples~\citep{geman1992neural, Neyshabur2015, spigler2018jamming,
  nakkiran2020deep}.
Despite recent progress in describing the implicit bias of stochastic gradient descent towards ``good'' minima~\citep{gunasekar2018characterizing, gunasekar2018implicit, Soudry2018, ji2019implicit, arora2019implicit, chizat2020implicit}, and the detailed analysis of solvable models of learning~\citep{ 
advani2020highdimensional, neal2018modern, mei2019generalization, belkin2019reconciling, hastie2022surprises, dascoli2020double, adlam2020understanding, lin2021causes, geiger2020scaling}, the mechanisms underlying this ``benign overfitting''~\citep{bartlett2020benign} in deep neural networks remain unclear, especially since their loss landscape contains ``bad'' local minima and SGD can reach them~\citep{liu2020bad}.

\begin{figure*}[!t]
  \centering
   \includegraphics[width=0.85\textwidth]{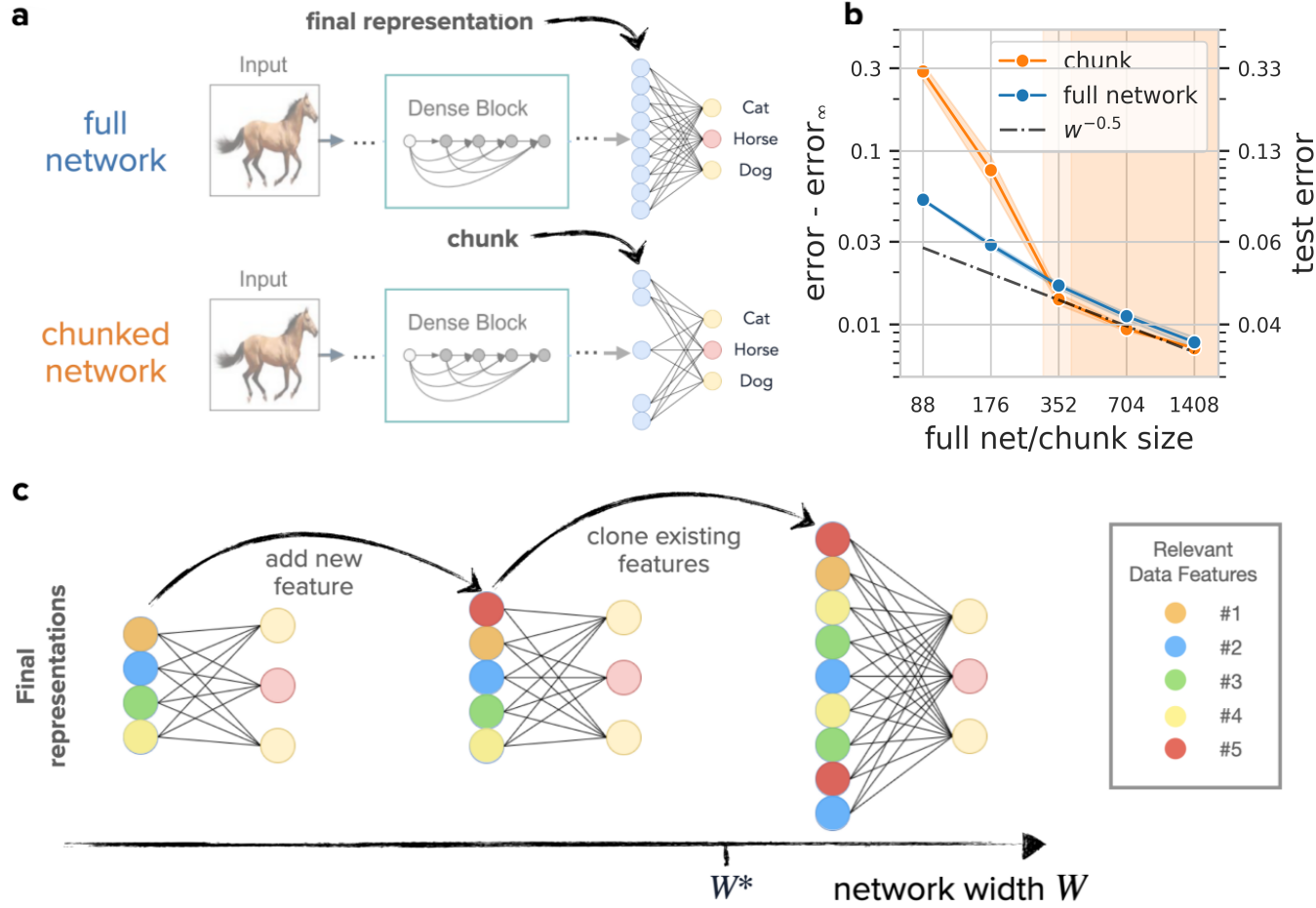}
  \caption{\label{fig:cartoon_} \textbf{The redundancy of representations in wide neural networks.} \textbf{a}: We analyze the final representations of
    deep neural networks (DNN),  namely the activities of the last hidden layer of neurons (light blue)
    We focus on the performance and the statistical properties of randomly chosen subsets of $w_c$ neurons which we call ``\emph{chunks}''. In the chunked network shown here, $w_c=5$ out of 9 neurons are kept and used to predict the output.
    \textbf{b}: 
    As we increase the size of the chunk $w_c$ that we keep in a state-of-the-art DNN, here a DenseNet40, the test error of the chunk (orange line) becomes similar to the test error of a full network of width $W=w_c$ (blue line). In this regime, which is reached when $w_c$ is larger than a threshold $w_c^*$ (shaded area) the error approaches its asymptotic value $\mathrm{error}_\infty$ as a power-law $w_c^{-1/2}$ (dashed line). $\mathrm{error}_\infty$ is the error of an ensemble average of 20 networks of the widest size.
    \textbf{c}: Illustration of three final representations for networks of increasing width. In small networks, an additional neuron fits new features of the data
    (red neuron). As the network width goes beyond a critical width $W^*$,
    additional neurons instead copy features already learned from data, instead of over-fitting to features that are not relevant to the task. This mechanism is
    suggested by the $w_c^{-1/2}$ decay of the chunk error, and by the
    statistical analysis, we present in this paper.}
\end{figure*}

In this paper, we describe a phenomenon in wide DNNs that could be a possible mechanism for benign overfitting when the networks are trained with regularization. We illustrate this mechanism in Fig.~\ref{fig:cartoon_} for a family of increasingly wide DenseNet40s~\citep{huang2017densely} trained on CIFAR10~\citep{krizhevsky2009learning} following common practice, in particular using weight decay (see Sec. \ref{sec:methods-nn}). 
For simplicity, we refer to the width $W$ of the last hidden representation as the width of the network. 
The blue line in Fig.~\ref{fig:cartoon_}-b shows that the average classification error
($\mathrm{error}$) approaches the performance of a large ensemble of  networks ($\mathrm{error}_\infty$) \citep{geiger2020scaling} as we increase the network width $W$. In agreement with \citep{zagoruyko2016wide}, we find that the performance of these DenseNets improves continuously with width. 
For widths greater than 350, the networks are wide enough to reach zero training error 
(see Appendix,  Sec. \ref{sec:additional_experiments}, Fig \ref{fig:app_densenet}-c) and, interestingly,  their test error decays approximately as $W^{\nicefrac{-1}{2}}$. Our goal is to understand how the error of the network can keep decaying beyond the interpolation threshold, and why it decays as $W^{\nicefrac{-1}{2}}$.

We make our key observation by performing the following experiment: we randomly
select a number $w_c$ of neurons from the last hidden layer of the widest DenseNet40 and remove all the other neurons from that layer as well as their connections (Fig. \ref{fig:cartoon_}-a). 
We then evaluate the performance of this chunk of $w_c$
neurons, \emph{without} retraining the network. The orange profile of Fig. \ref{fig:cartoon_}-b shows the test error of \emph{chunks} of varying
sizes.
There are two regimes: for small chunks, the error decays faster than $w_c^{\nicefrac{-1}{2}}$, while beyond a critical chunk size $w_c^*$ (shaded area), the error of a chunk of $w_c$ neurons is roughly the same as the one of a full network with
$w_c$ neurons. Furthermore, the error of the chunks decays with the same
power-law $w_c^{\nicefrac{-1}{2}}$ beyond this critical chunk size.

The decay rate of $\nicefrac{-1}{2}$ suggests that in this regime chunks of  $w_c$ neurons can be thought of as statistically independent estimators of the same features of the data, differing only by small, uncorrelated noise. 
In other words, beyond the critical width
$w_c^*$, the final hidden representation of an input in a trained, wide DNN becomes highly redundant.
This motivates a possible mechanism for benign overfitting, schematically portrayed in Fig. \ref{fig:cartoon_}-c: as the network
becomes wider, additional neurons are first used to learn new features of the data. 
Beyond the critical width~$w_c^*$, additional neurons in the final layer don't fit new features in the data, and hence over-fit; instead, they make a copy, or a \emph{clone}, of a feature that is already part of the final representation. The last layer thus splits into more and more clones as the network grows wider as we illustrate at the bottom of
Fig.~\ref{fig:cartoon_}. The accuracy of these wide networks then improves with
their width because the network implicitly averages over an increasing number of clones in its representations to make its prediction. 

This paper provides a quantitative analysis of this phenomenon on various data sets and architectures. Our main findings can be summarized as follows:

\begin{enumerate}
\item A chunk of $w_c$ random neurons of the last hidden representation of a wide neural network predicts the output with an error that decays as $w_c^{-1/2}$ if the layer is wide enough and $w_c$ is large enough. In this regime, we call the chunk a ``clone'';
\item  Clones fit the training set with zero error and can be linearly mapped one to another, or to the full representation, with an error that can be described as uncorrelated random noise.
\item Clones appear if the model is trained with weight decay and the training set is fitted with zero error.
If training is stopped too early or if the training is performed without regularization, 1.~and 2.~do not take place, even if the last representation is very wide. 
\end{enumerate}

\section{Methods}
\label{sec:methods}

\subsection{Neural network architectures}
\label{sec:methods-nn}

We report experimental results obtained with several architectures (fully
connected networks, Wide-ResNet-28, DenseNet40, ResNet50) and data sets
(CIFAR10/100~\cite{krizhevsky2009learning},
ImageNet~\cite{imagenet_cvpr09}). 
We train all the networks using SGD with momentum and, importantly, weight decay. The amount of weight decay is found with a small grid search, while the other relevant hyperparameters are set following standard practice. 
We give detailed information on our training setups in Sec.  \ref{sec:training_hyperparams} of the Appendix. All our experiments are run on Volta V100 GPUs. In the following paragraphs, we describe how we vary the width $W$ of the models.

\paragraph{Fully-connected networks on MNIST.}
We train a fully-connected network to classify the parity of the MNIST
digits~\citep{lecun1998} (pMNIST) following the protocol of Geiger \emph{et al.}~\cite{geiger2020scaling}.  MNIST digits are projected on the first ten principal
components, which are then used as inputs of a five-layer fully-connected
network (FC5).
The four hidden representations have the same width $W$ and the output is a real number whose sign is the predictor of the parity of the input digit.

\paragraph{Wide-ResNet-28 and DenseNet40 on CIFAR10/100.} We train CIFAR10 and CIFAR100 on family of
Wide-ResNet-28~\citep{zagoruyko2016wide} (WR28). 
The number $W$ of the last hidden neurons in a WR28-$n$ is $64 \cdot n$, obtained after average pooling the last $64 \cdot n$ channels of the network. In our experiments, we also analyze two narrow versions of the standard WR28-1 which are not typically used in the literature. We name them  WR28-0.25 and WR28-0.5 since they have 1/4 and 1/2 of the number of channels of WR28-1.
Our implementation of DenseNet40 follows the DenseNet40-BC variant \citep{huang2017densely}. 
We vary the number of input channels $c$  in $\{16, 32, 64, 128, 256\}$, which is twice the growth rates $k$ of the networks \cite{huang2017densely}. The number $W$ of the last hidden features of this architecture is $5.5\cdot c$.

\paragraph{ResNet50 on ImageNet.} We modify the ResNet50 architecture~\citep{he2016deep} by multiplying by a constant factor $c \in \{0.25, 0.5, 1, 2, 4\}$ the number of channels of all the layers after the input stem. 
When $c = 2$ our networks differ from the standard Wide-ResNet50-2~\cite{zagoruyko2016wide} since we double the channels of \emph{all} the layers and not just those of the bottleneck of the ResNet blocks.
%
As a consequence in our implementation, the number of features after the last pooling layer is $W = 2048 \cdot c$ while in~\cite{zagoruyko2016wide} $W$ is fixed to 2048. 

\subsection{Analytical methods}
\label{sec:methods-analysis}

\paragraph{Reconstructing the wide representation from a smaller chunk.}
To determine how well a subset of $w$ neurons can reconstruct the full representation of size $W$ we search for the $W \times w$ linear map~$\mathbf{A}$, able to minimize the squared difference 
${({\mathbf{x}}^{(W)}-\hat{\mathbf{x}}^{(W)})}^2$ between the $W$ activations of the full layer representation, $\mathbf{x}^{(W)}$, and the activations predicted from the chunk of size $w$, $\hat{\mathbf{x}}^{(W)}$:
\begin{equation}
    \label{eq:fit}
    \hat{\mathbf{x}}^{(W)} = \mathbf{A} \mathbf{x}^{(w)}.
\end{equation}
%
This least-squares problem is solved with ridge regression \citep{hastie01statisticallearning} with regularization set to $10^{-8}$, and we use the $R^2$ coefficient of the fit to measure the predictive power of a given chunk size.
The $R^2$ value is computed as an average of the  single-activations $R^2$ values corresponding to the $W$ output coordinates of the fit, weighted by the variance of each coordinate.
We further compute the $W \times W$ covariance matrix $C_{ij}$ of the \emph{residuals} of this fit, and from $C_{ij}$ we obtain the correlation matrix as:
\begin{equation}
    \label{eq:correlation}
    \rho_{ij} = \frac{C_{ij}}{\sqrt{C_{ii}C_{jj}} + 10^{-8}},
\end{equation}
with a small regularization in the denominator to avoid instabilities when the standard deviation of the residuals falls below machine precision.
To quantify how much the errors of the fit are correlated, we average the absolute values of the non-diagonal entries of the correlation matrix $\rho_{ij}$.
For short, we refer to this quantity as a `mean correlation'.

\paragraph{Reproducibility.} We provide code to reproduce our experiments and our
analysis online
at~\url{https://github.com/diegodoimo/redundant_representation}.

\section{Results}
\label{sec:results}
\begin{wrapfigure}{r}{0.4\textwidth}
  \vspace{-1cm}
  \centering
  \includegraphics[width=0.45\textwidth]{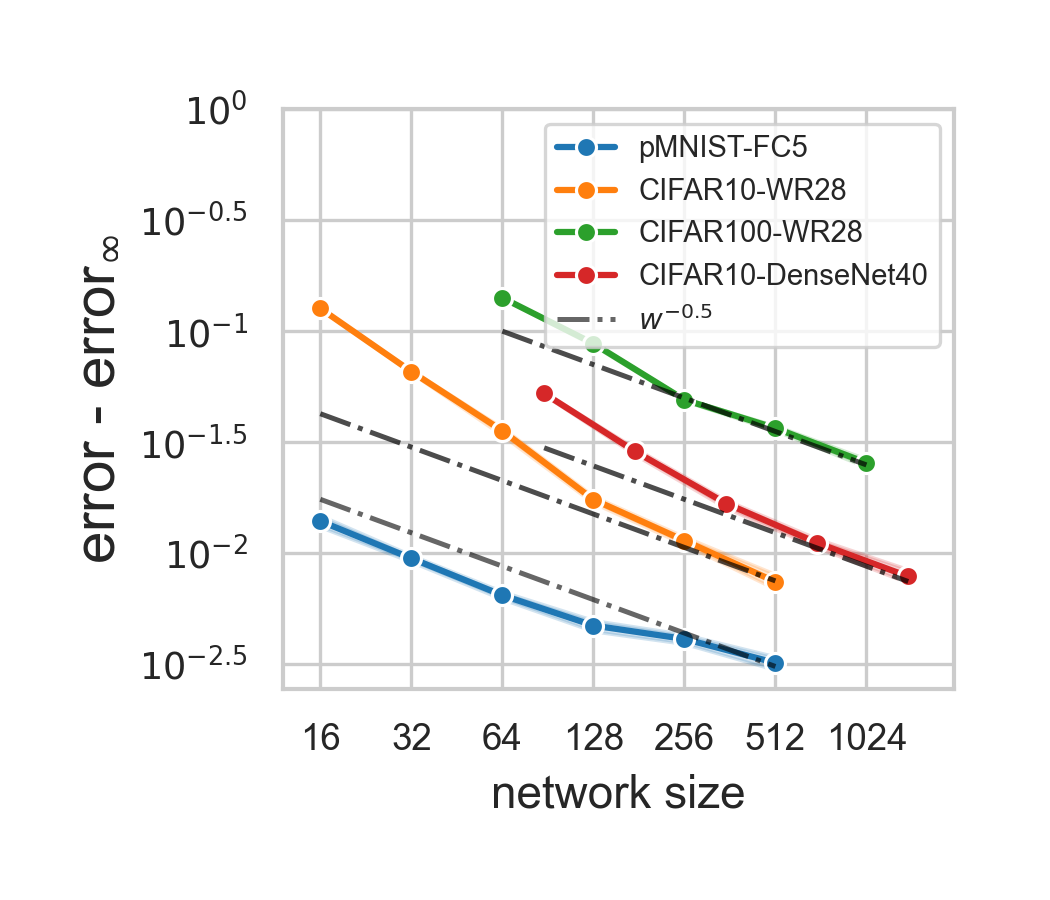}
  \vspace{-0.8cm}
\caption{\label{fig:error-full-network} 
      \textbf{Scaling of the test error with
      width for various DNN.} The average test error of neural networks with various architectures approaches the test error of an ensemble of such networks as the network width increases. The network size shown here is the width of the final representation. For large width, we find a power-law behavior $\mathrm{error} - \mathrm{error}_\infty \propto W^{-1/2}$ across data sets and architectures. Full experimental details in Sec.~\ref{sec:methods-nn}}
\vspace{-0.5cm}
\end{wrapfigure}
  
\paragraph{The test error of chunks of $w_c$ neurons of the final representation asymptotically scales as~$w_c^{-1/2}$.}
 The mechanism we propose is inspired by the following experiment: we compute the test accuracy of models obtained by selecting a random subset of $w_c$ neurons from the \emph{final hidden representation} of a wide neural network. 
 We select $w_c$ neurons at random  and we compute the test accuracy of a network in which we set to zero the activation of all the other $w-w_c$ neurons of the final layer. 
Importantly, we do not fine-tune the  weights  after selecting  the $w_c$ neurons: all the remaining parameters of the previous layers are left unchanged and only the the activations of the "killed" neurons of the last hidden representation are not used to compute the logits.
We take 500 random samples of neurons for each chunk width $w_c$.
 We consider three different data sets: pMNIST trained on a fully connected network, CIFAR10 and CIFAR100 trained on convolutional networks. 
 The width $W$ of the network is 512 for pMNIST and CIFAR10,  and $W=1024$ for CIFAR100 (see Sec. \ref{sec:methods-nn}).
 In all these cases, $W$ is large enough to be firmly in the regime where the accuracy of the networks scales (approximately) as $W^{-1/2}$ (see Fig.~\ref{fig:error-full-network}). 
 %

\begin{figure*}[!t]
  \centering
  \includegraphics[width=1.\textwidth]{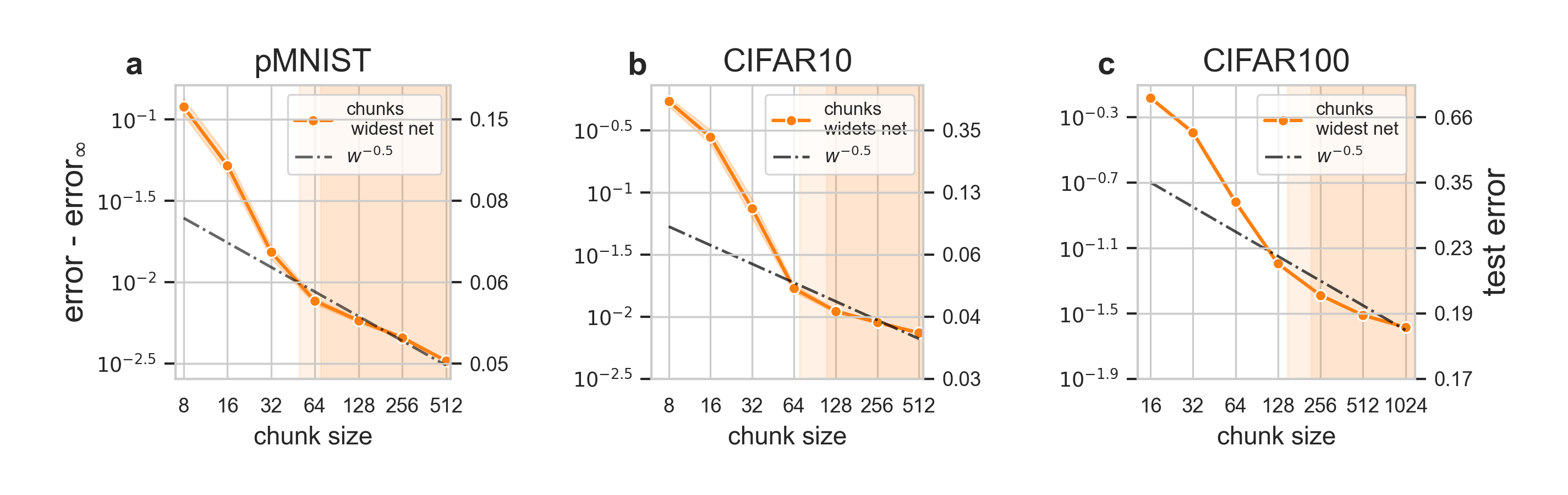}
  \caption{\label{fig:chunk-error-scaling} \textbf{Scaling of the test error of chunks of neurons extracted from the final hidden  representation of wide NNs.} We plot how the test error of chunked networks approaches $\mathrm{error}_\infty$, the error of an ensemble of 20 networks of the widest size (e.g. $W = 1024$ for CIFAR100), as the chunk size $w_c$ increases. Chunks are formed by selecting a number of $w_c$ neurons at random from the final hidden representation of the widest networks: a FC5 on pMNIST (width $W=512$), and Wide-ResNet-28 for CIFAR10 ($W=512$) and CIFAR100 ($W=1024$). The shaded regions indicate regions where the error of the chunks with $w_c$ neurons decays as $w_c^{-1/2}$.}
\end{figure*}

In  Fig.~\ref{fig:chunk-error-scaling} we plot the test error of the "chunked models" as a function of $w_c$ (orange lines). The behavior is similar in all three networks: the test error decays as  $w_c^{\nicefrac{-1}{2}}$ for chunks that are larger than a critical value $w_c^*$, which depends on the data set and architecture used.  
This decay follows the same law observed for full networks of the same width  (Fig.~\ref{fig:error-full-network}). This implies that a model obtained by selecting a random chunk of $w_c > w_c^*$ neurons from a wide final representation behaves similarly to a full network of width $W=w_c$. 
Furthermore, a decay with rate~$\nicefrac{-1}{2}$ suggests that the final representation of the wide networks can be thought of as a collection of statistically independent estimates of a finite set of data features relevant for classification. Adding neurons to the chunk hence reduces their prediction error in the same way an additional measurement reduces the measurement uncertainty, leading to the~$\nicefrac{-1}{2}$ decay.

\vspace{-0.0cm}

At  $w_c$ smaller than $w_c^*$ instead, the test error of the chunked models decays faster than $w_c^{\nicefrac{-1}{2}}$ in all the cases we considered, including the DenseNet architecture trained on CIFAR10 shown in Fig.~\ref{fig:cartoon_}-b.
In this regime, adding neurons to the final representation improves the quality of the model significantly quicker than it would in independently trained models of the same width (see Fig. \ref{fig:cartoon_}-c for a pictorial representation of this process).
We call chunks of neurons of size  $w_c \geq w_c^*$ \emph{clones}. 
%
In the following, we characterize more precisely the properties of the clones. 

\begin{figure*}
    \centering
    \includegraphics[width=1.\textwidth]{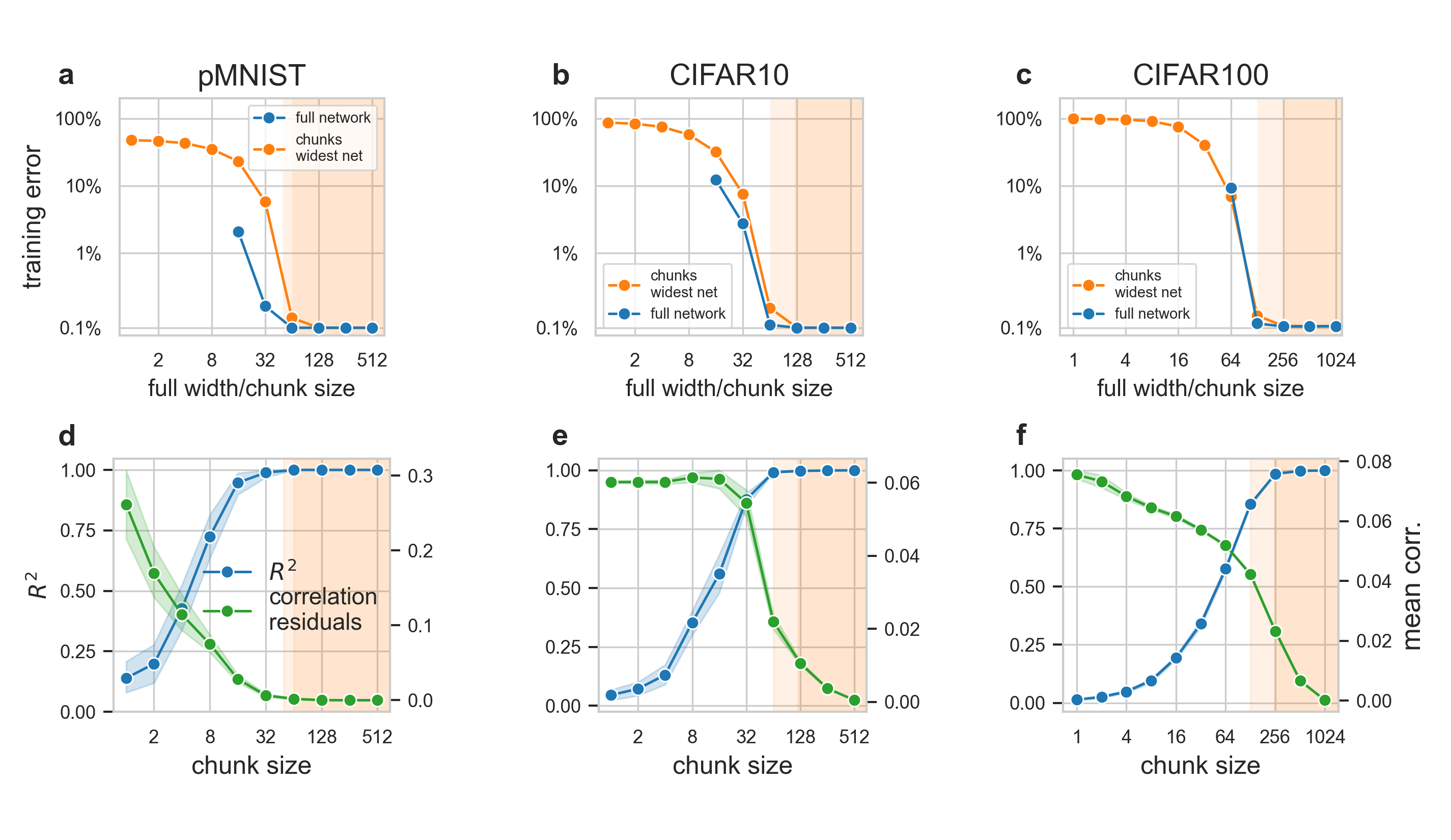}
    \caption{\label{fig:hallmarks-redundancy}\textbf{The three signatures of representation redundancy.} \textbf{(i)} The training errors of the full networks (blue) and of the chunks taken from the widest network (orange) approach zero beyond a critical width/chunk size, resp. (panels {\bf a}-{\bf c}). \textbf{(ii)} The final representation of the widest network can be reconstructed from a chunk using linear regression~\eqref{eq:fit} with an explained variance $R^2$ close to 1 (blue lines in panels {\bf d}-{\bf f}). \textbf{(iii)} The residuals of the linear map can be modeled as independent noise: we show this by plotting the mean correlation of these residuals (green line, panels {\bf d}-{\bf f}), averaged over 100 reconstructions starting from different chunks. A low correlation at high $R^2$ indicates that the chunk contains the information of the full representation with some statistically independent noise. \emph{Experimental setup:} FC5 on pMNIST, Wide ResNet-28 on CIFAR10/100. Full details in Methods section~\ref{sec:methods-nn}}
\end{figure*}

\paragraph{Clones interpolate the training data.}
A trained deep network often represents the salient features of the data set well enough to  achieve (close to) zero classification error on the training data. 
In the top panels of Fig.~\ref{fig:hallmarks-redundancy}, we show that wide networks can interpolate their training set also using just a subset of $w_c > w_c^*$ random neurons: the dark orange profiles show that when the size of a chunk is greater than $w_c^* \sim$ 50 for pMNIST, 100 for CIFAR10 and 200 for CIFAR100, the predictive accuracy on the training set remains almost 100\%.
The minimal size of a clone $w_c^*$ can be identified with the minimal number of neurons required to interpolate the training set.
Beyond $w_c^*$, the neurons of the final representation become redundant since the training error remains (close to) zero even after removing neurons from it. 
The number of distinct clones in a network of width $W$ is $n = {{W}/{w_c^*}}$. If  distinct clones provide independent measures of the same salient features of the data, the \emph{test} error decays approximately as $n^{\nicefrac{-1}{2}}$ or equivalently $W^{\nicefrac{-1}{2}}$. 
In the following, we will indeed see that distinct clones differ from each other by uncorrelated random noise.

\paragraph{Clones reconstruct the full representation almost perfectly.} From a geometrical perspective, the important features of the final representation correspond to directions in which the data landscape shows large variations \citep{bengio2013representation}.
A clone is a chunk that is wide enough to encode almost exactly these directions 
(since its training error is almost zero), 
but using much fewer neurons than the full final representation.
We analyze this aspect by performing a linear reconstruction of the $W$ activations of the last hidden representation of the widest network starting from a random subset of $w_c$ activations using ridge regression with a small regularization penalty according to Eq.~\eqref{eq:fit}.
The  blue profiles in Fig.~\ref{fig:hallmarks-redundancy}-(d,e,f),  show the $R^2$ coefficient of fit as a function of the chunk size $w_c$ for pMNIST (left), CIFAR10 (center), CIFAR100 (right).
When $w_c$ is very small, say below $6$ for pMNIST, $20$ for CIFAR10 and $60$ for CIFAR100, the~$R^2$ coefficient grows almost linearly with $w_c$ \footnote{The linear trend can not be clearly seen in Fig. \ref{fig:hallmarks-redundancy} as we plot the $x$-axis with a logarithmic scale.}. In this regime, adding a randomly chosen activation from the full representation to the chunk increases substantially  $R^2$.
When $w_c$ becomes larger  $R^2$ reaches almost one. 
This transition happens when $w_c$ is still much smaller than $W$  and corresponds approximately to the regime in which the test error starts scaling with the inverse square root of $w_c$ (see Fig.~\ref{fig:chunk-error-scaling}).  
The almost perfect reconstruction of the original data landscape with few
neurons is a consequence of the low \emph{intrinsic dimension} (ID) of
the representation~\citep{ansuini2019intrinsic}.
The ID of the widest representations gives a lower bound on the number of coordinates required to describe the data manifold, and hence on the neurons that a chunk needs in order to have the same classification accuracy as the whole representation. The ID of the last hidden representation is 2 in pMNIST, 12 in CIFAR10, 14 in CIFAR100, numbers which are much lower than $w_c^*$, the width at which a chunk can be considered a clone.

\paragraph{Clones differ from each other by uncorrelated random noise.} 
When~$w_c > w_c^*$ the small residual difference between the chunked representation and the full representation can be approximately described as  statistically independent random noise.
The green profiles of the bottom panels of Fig. \ref{fig:hallmarks-redundancy} show
the \emph{mean correlation} of the residuals of the linear fit (see Sec. \ref{sec:methods-analysis}).
%
Below $w_c^*$, the residuals are not only large but also significantly correlated, since they are related to relevant features of the data that are not covered by the neurons of the chunk. As the chunk width increases above  $w_c^*$, the correlation between residuals drops basically to zero.
Therefore, in networks wider than $w_c^*$ any two chunks of equal size $w_c > w_c^*$ can be effectively considered as equivalent copies, or clones, of the same representation (that of the full layer), differing only by a small and non-correlated noise, consistently with the scaling law of the error shown in Fig.~\ref{fig:chunk-error-scaling}.
\vspace{-0cm}

\paragraph{The dynamics of training.}
In the previous paragraphs, we set forth evidence in support of the hypothesis
that large chunks of the final representation of wide DNNs behave approximately
like an ensemble of independent measures of the full feature space. This allowed
us to interpret the decay of the test error of the full networks with the 
network width observed empirically in Fig.~\ref{fig:error-full-network}. The
three conditions that a chunked model satisfies in the regime in which its test
error decays as $w_c^{\nicefrac{-1}{2}}$ are represented in
Fig.~\ref{fig:hallmarks-redundancy}: (i) the training error of the chunked model is close to zero; (ii) the chunked model can be used to reconstruct the full final representation with an
$R^2\sim 1$ and (iii) the residuals of this reconstruction can be modeled as
independent random noise. These three conditions are all observed at the end of
the training.
We now analyze the training \emph{dynamics}. We will see that for the clones to arise, 
models not only need to be wide enough but also, crucially, they need to be trained to maximize their performance.

\begin{figure*}
\centering
\includegraphics[width=1.\textwidth]{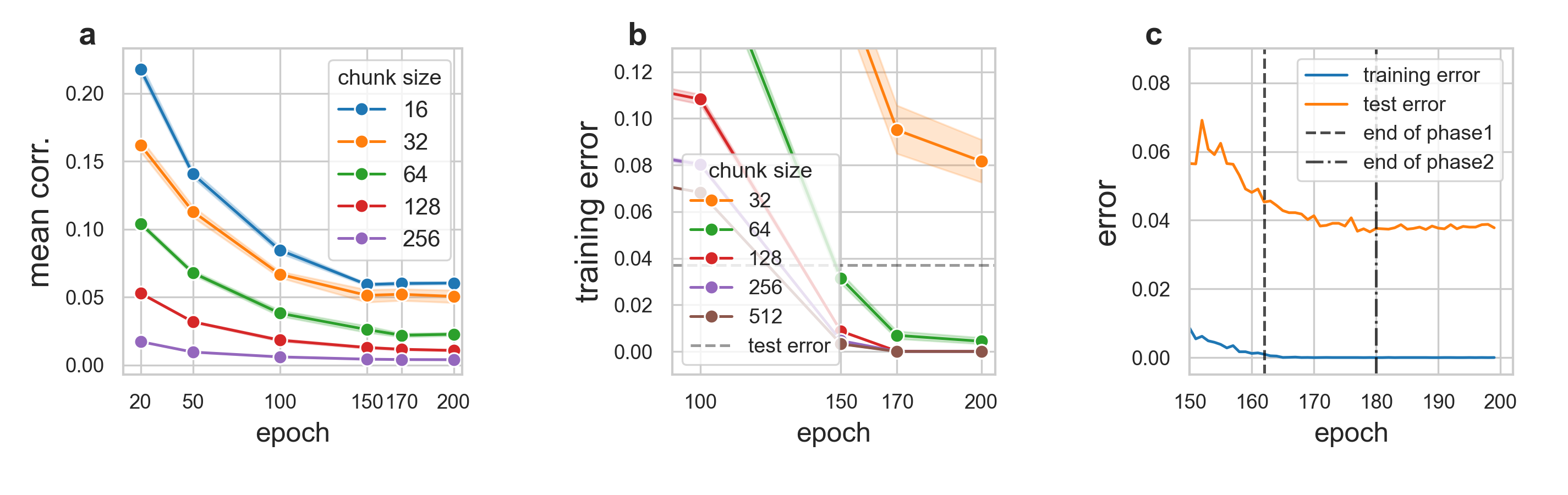}
\caption{
\textbf{The onset of clones during training.}
\textbf{a:} As in Fig.~\ref{fig:hallmarks-redundancy}, we show the mean correlation of the residuals of the linear reconstruction of the final representation from chunks, but this time as a function of training epochs. A small correlation indicates that the reconstruction error in going from chunks to final representation can be modeled as independent noise. Data obtained from the same WR28-8 trained on CIFAR10 as in Fig.~\ref{fig:hallmarks-redundancy}.
\textbf{b:} Training error during training for chunks of different sizes.
After the network has reached zero training error at $\sim 160$ epochs, continuing to train improves the training accuracy of the chunks.
\textbf{c:} Test and training  error during training for the full network. Between epochs 160 and 180, the clones of the full network progressively achieve zero training error. In the same epochs, one observes  a small improvement in the test error.
}
\label{fig:training_dynamics}
\end{figure*}

Clones are formed in two stages, which occur at different times during training.
The first phase begins as soon as training starts: the network gradually adjusts the chunk representations in order to produce independent copies of the data manifold.
This can be clearly observed in Fig.~\ref{fig:training_dynamics}-a, which depicts the mean correlation between the residuals of the linear fit from the chunked to the full final representations of the network, the same quantity that we analyze in Fig.~\ref{fig:hallmarks-redundancy}-(d-e-f, green profiles), but now as a function of the training epoch. Both Figs.~\ref{fig:hallmarks-redundancy} and~\ref{fig:training_dynamics} analyze the WR28-8 on CIFAR10.
As training proceeds, the correlations between residuals diminish gradually until epoch 160 and become particularly low for chunks greater than 64. After epoch 160 further training does not bring any sizeable reduction in their correlation.
At epoch 160 the full network also achieves zero error on the training set, as shown in  Fig.~\ref{fig:training_dynamics}-b (brown) and Fig.~\ref{fig:training_dynamics}-c (blue).
This event marks the end of the first phase and the beginning of the second phase where the training error of the clones keeps decreasing while the full representation (blue) has already reached zero training error.
For example, chunks of size 64 at epoch 150 have training errors comparable to the test error (dashed line of the middle panel).
In the subsequent $\sim 20$  epochs the training error of clones of size 128 and 256 reaches exactly zero, and the training error of chunks of size 64 reaches a plateau. 

Importantly, both phases improve the generalization properties of the network.
This can be seen in Fig.~\ref{fig:training_dynamics}-c, which reports the training and test error of the network, with the two phases highlighted.
The figure shows that both phases lead to a reduction in the test error, although the first phase leads by far to the greatest reduction, consistent with the fact that the greatest improvements in accuracy typically arise during the first epochs of training.
The formation of clones can be considered finished around epoch 180 when all the clones have reached almost zero error on the training set. After epoch 180 we also observe that the test error stops improving.
In the Appendix (Sec. \ref{sec:additional_experiments}) we report the same analysis done on CIFAR100 (see Fig. \ref{fig:app_cifar100_dyn}) and CIFAR10 trained on a DenseNet40 (see Fig. \ref{fig:app_densenet}-(d-e-f)).

\begin{figure}
\centering
\includegraphics[width=0.7\columnwidth]{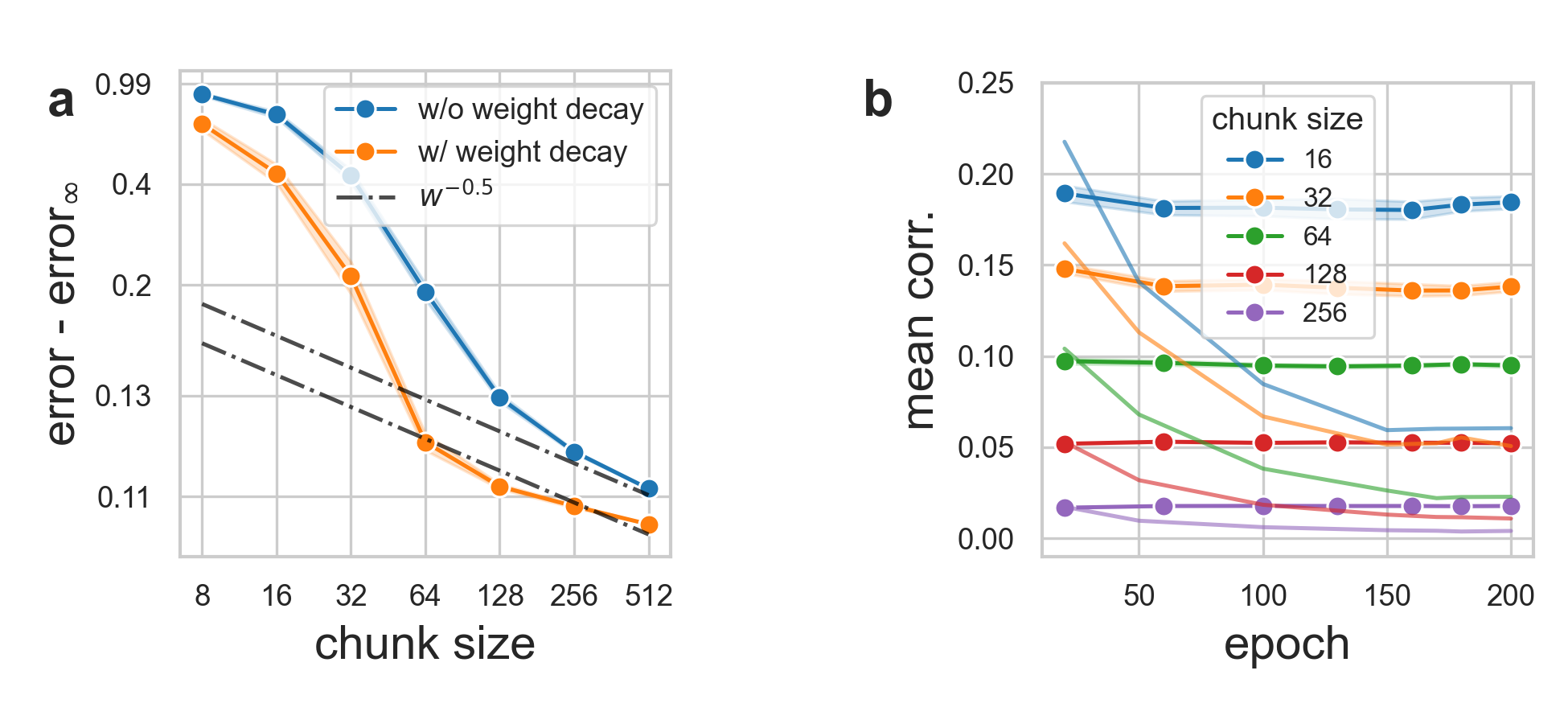}
\caption{\label{fig:narrow_bad-trained} \textbf{A network trained without weight decay on CIFAR10.} \textbf{a:} the test error of chunks of a Wide-ResNet28-8 trained without weight decay (blue) and with weight decay (orange, taken from Figure \ref{fig:chunk-error-scaling}-b). \textbf{b:}  Mean correlation between residuals of the linear reconstruction of the full representation from chunks of different sizes for two networks: one trained without weight decay (thick lines), and one using weight decay (thin lines, same data as in Fig.~\ref{fig:training_dynamics}-a).
}
\end{figure}

\paragraph{Clones appear only in regularized networks.}
So far in this work, we have shown only examples of regularized networks and data sets in which representations are redundant. 
However, if the network is not regularized, some of the signatures described above don't appear even if the width of the final representation is much larger than $W^*$ (the minimum interpolating width). 
Figure \ref{fig:narrow_bad-trained} shows the case of the Wide-ResNet28-8 analyzed in Fig. \ref{fig:training_dynamics} trained on exactly the same data set (CIFAR10) but
without weight decay.
As shown in Fig.\ref{fig:narrow_bad-trained}-a in the network trained without regularization (blue line) the error does not scale as $w_c^{-1/2}$. 
This, as we have seen, indicates that the last hidden representation cannot be split in clones equivalent to the full layer.
Indeed, the mean correlation of the residuals of the linear map of the chunks to the full representation remains approximately constant during training (Fig. \ref{fig:narrow_bad-trained}-b), and is always much higher than what we observed for the same architecture and data set when training is performed with weight decay.
We performed a similar analysis on the DenseNet40 (see Fig. \ref{fig:app_densenet_not_reg}), observing an analogous trend.

\begin{figure}[!b]
  \centering
  \includegraphics[width=\textwidth]{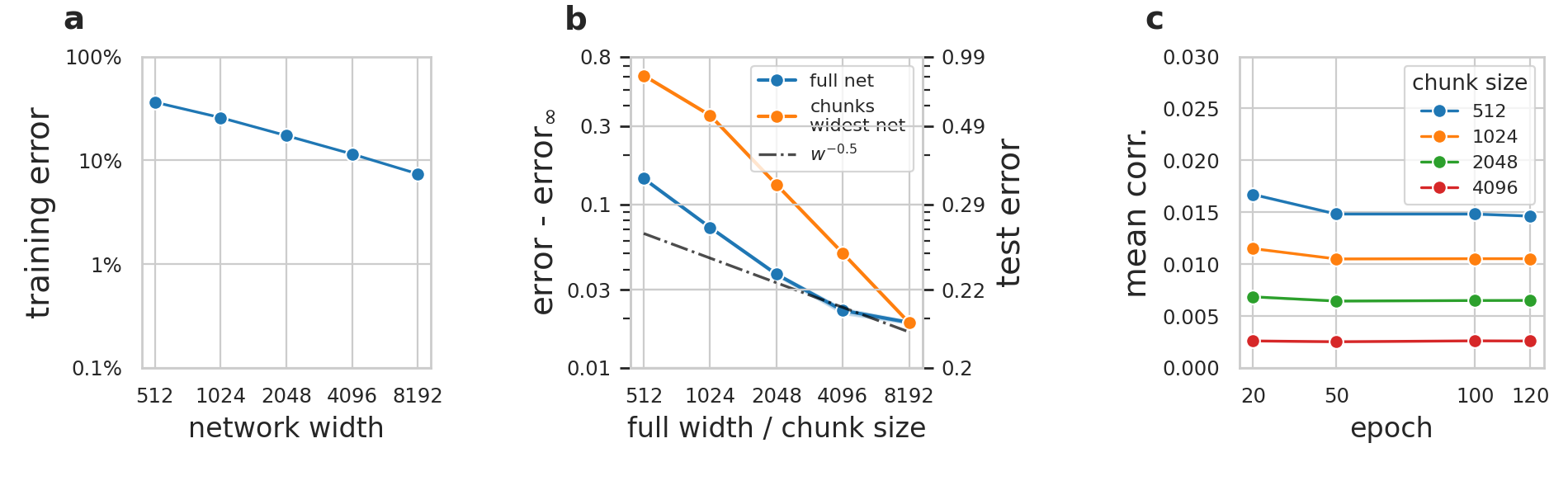}
  \caption{\label{fig:imagenet}\textbf{ResNet50 trained on ImageNet}
\textbf{a:} ImageNet training error as a function of the ResNet50 width.
  \textbf{b:} Decay of the test error as a function of the network width (blue) and for chunks of the widest ResNet50 (orange) to the error of an ensemble of ResNet50-4.  The ensemble consists of four networks.
  \textbf{c:}
  Mean correlation (see Sec. \ref{sec:methods-analysis}) of the residuals of the linear map of a chunk of the last hidden representation to the full representation. The network analyzed is ResNet50-4.
 }
\end{figure}

\paragraph{Clones appear only if a network interpolates the training set: the case of ImageNet.} %
We saw that a chunk of neurons can be considered a clone if it fully captures the relevant features of the data, achieving almost zero training error (see Fig. \ref{fig:hallmarks-redundancy}). 
This condition is not satisfied for most of the networks trained on ImageNet \cite{belkin2019reconciling}, therefore we do not expect to see redundant representations in this important case.
We verified this hypothesis by training a family of ResNet50s where we multiply all the channels of the layers after the input stem by a constant factor $c \in \{0.25, 0.5, 1, 2, 4\}$. In this manner the widest final representation we consider consists of $8192$ neurons, which is four times wider than both the standard ResNet50 \citep{he2016deep} and its wider version \citep{zagoruyko2016wide} (see Sec. \ref{sec:methods-nn}). 
We trained all the networks following the standard protocols and achieved test errors comparable to or slightly lower than those reported in the literature (see Appendix, Sec.  \ref{sec:training_hyperparams}).
We find that even in the case of the largest ResNet50, the top-1 error on the training set is $\sim$ 8\% (see Fig. \ref{fig:imagenet}-a) and the network does not achieve interpolation, as discussed also in \cite{belkin2019reconciling}.

In this setting, none of the elements associated with the development of independent clones can be observed. The scaling of the test error of the chunks is steeper than $w_c^{-1/2}$ (see Fig. \ref{fig:imagenet}-b) suggesting that chunks remain significantly correlated to each other. Figure \ref{fig:imagenet}-c shows that the mean correlation of the residuals does not decrease during training, as it happens for the networks we trained on CIFAR10 and CIFAR100.
We conclude that in a ResNet50, a representation with 8192 neurons is too narrow to encode all the relevant features redundantly on ImageNet, and a chunk as large as 4096 activations is not able to reconstruct all the relevant variations of the data as it does in the cases analyzed in Sec. \ref{sec:results}.

\section{Discussion}
\label{sec:discussion}
This work is an attempt to explain the paradoxical observation that over-parameterization boosts the performance of DNNs.
This ``paradox'' is  not a peculiarity of DNNs: if one trains a prediction model with $n$ parameters using the same training set, but starting from independent initial weights and receiving samples in an independent way, one can obtain, say, $m$ models which, in suitable conditions, provide  predictions of the same quantity with independent noise due to initialization, SGD schedule, etc.
If one estimates the target quantity by an ensemble average, the statistical error  will (ideally) scale with $m^{\nicefrac{-1}{2}}$, and therefore with $N^{\nicefrac{-1}{2}}$, where $N=n\,m$ is the total number of parameters of the combined model.
This will happen even if $N$ is much larger than the number of data.

What is less trivial is that a DNN can accomplish this scaling within a single model, in which all the parameters are optimized collectively via the minimization of a single loss function.
Our work describes a possible mechanism at the basis of this phenomenon in the special case of neural networks in which the last layer is very wide and the model is regularized.
We observe that if the layer is wide enough, random subsets of its neurons can be viewed as approximately independent representations of the same data manifold (or clones).
%
%
This implies a scaling of the error with the width of the layer as~$W^{\nicefrac{-1}{2}}$, which is qualitatively consistent with our observations. 

\paragraph{The impact of network architecture.}
The capability of a network to produce statistically independent clones is a genuine effect of the over-parametrization of the \emph{whole} network as we find that redundancies appear even if the last layer width is kept constant and the width of all intermediate layers is increased (see Appendix, Sec.~\ref{ref:app_ony_last}, Fig.~\ref{fig:app_only_last}-a, ). 
At the same time, we also verified that if the network is too narrow to interpolate the training set, increasing the width of \emph{only} the final representation is not sufficient to make the last layer redundant. We give an example of this effect in Fig.~\ref{fig:app_only_last}-b, where we show that the test error of a WR28-1 on CIFAR10 does not decrease if only the width of the final representation is increased, while the rest of the architecture is kept at a constant width.

\paragraph{The impact of training.}
The mechanism we described is robust to different training objectives 
since we trained the convolutional networks with cross-entropy loss and the fully connected networks with hinge loss.
However, even for wide enough architectures, clones appear only if the training is continued until the training error reaches zero. 
In our examples, by stopping the training too early, for example when the training error is similar to the test error, the chunks of the last representation would not become entirely independent from one another, and therefore they could not be considered clones.

\paragraph{Neural scaling laws.} Capturing the asymptotic performance of neural networks via scaling laws is an active research area. Hestness \emph{et al.}~\citep{hestness2017deep} gave an experimental analysis of scaling laws w.r.t.~the training data set size in a variety of domains. Rosenfeld \emph{et al.} and Kaplan \emph{et al.}~\citep{rosenfeld2020constructive, kaplan2020scaling} experimentally explored the scaling of the generalization error of deep networks with the number of parameters/data points across architectures and application domains for supervised learning, while Henighan \emph{et al.}\citep{henighan2020scaling}
identified empirical scaling laws in generative models.
Bahri \emph{et al.}~\citep{bahri2021explaining} showed the existence of four scaling regimes and described them theoretically in the NTK or \emph{lazy regime}~\cite{jacot2018neural,du2018gradient,chizat2019lazy}, where the network weights stay close to their initial values throughout training. None of these works propose a mechanism that would explain these scalings with properties of the representation.
%
%
Geiger \emph{et al.} found that the generalization error can be related to the fluctuations of the output induced by initialization and showed that it scales as $W^{-1}$ in networks trained \emph{without} weight decay both in the NTK \citep{geiger2020scaling} and in the mean field \citep{geiger2020disentangling} regimes.
We instead consider the feature learning regime and train our networks with weight decay which is unavoidable to obtain models with state-of-the-art performance. This might explain the difference in the scaling law that we observe empirically. 
%
%
Previous theoretical work 
did not study the impact of weight decay on scaling laws, so we hope that our results can spark further studies on the role of this essential regularizer.

%

\paragraph{Relation to theoretical results in the mean-field regime.}
Our empirical results also agree with recent theoretical results that were obtained for two-layer neural networks~\citep{Mei2018, rotskoff2018interacting, Chizat2018, Sirignano2018, goldt2019dynamics, refinetti2021classifying}.
These works characterize the optimal solutions of
two-layer networks trained on synthetic data sets with some controlled features.
In the limit of infinite training data, these optimal solutions correspond to
networks where neurons in the hidden layer duplicate the key features of the
data.
These ``denoising solutions'' or ``distributional fixed points'' were found for networks with wide hidden layers~\citep{Mei2018, rotskoff2018interacting, Chizat2018, Sirignano2018} and wide input
dimension~\citep{goldt2019dynamics, refinetti2021classifying}. Another point of connection
with the theoretical literature is the concept of \emph{dropout stability}. A network is said to be $\epsilon$-dropout stable if its training loss changes by less than $\epsilon$ when 
half the neurons are removed at random from each of its layers~\cite{kuditipudi2019explaining}. Dropout stability has been rigorously linked to several phenomena in neural networks, such as the connectedness of the minima of their training landscape~\cite{shevchenko2020landscape, nguyen2021connectivity}.

\paragraph{Bias-variance trade-off and implicit ensembling} The success of various deep
learning architectures and techniques has been linked to some form of ensembling. 
The successful dropout regularisation technique~\citep{hinton2012improving, srivastava2014dropout}
samples from an exponential number of ``thinned'' networks during training to
prevent co-adaptation of hidden units. While this can be seen as a form of
(implicit) ensembling, here we observe that co-adaptation of
hidden units in the form of clones occurs \emph{without} dropout, and is crucial for their improving performance
with width. Recent theoretical work on random features showed that ensembling and
over-parameterization are two sides of the same coin and that both mitigate the increase in the variance of the network that classically leads to \emph{worse} performance with over-parameterization due to the bias-variance trade-off~\citep{dascoli2020double,
  adlam2020understanding, lin2021causes}.
  The plots of bias and variance in Fig.~\ref{fig:app_bias_variance_ls0} for the architectures trained on the CIFAR10 and CIFAR100 data sets show that the clone size
  in these cases is slightly above the peak of the variance and almost coincides with the interpolation width of the full networks of the same size.
  

{\bf Impact for applications.} The framework introduced in this work allows verifying if a neural network is sufficiently expressive to encode multiple statistically independent representations of the same ground truth, which, we believe, is a fair proxy of model quality and robustness. 
In particular, we find that reaching interpolation on the training set is not necessarily detrimental for generalization, and is instead a necessary condition for developing redundancies which, in turn, reduces the test error.  

\bibliography{ensembling}

\begin{thebibliography}{10}

\bibitem{zhang2016understanding}
C.~Zhang, S.~Bengio, M.~Hardt, B.~Recht, and O.~Vinyals.
\newblock {Understanding deep learning requires rethinking generalization}.
\newblock In {\em ICLR}, 2017.

\bibitem{arpit2017closer}
D.~Arpit, S.~Jastrzebski, M.S. Kanwal, T.~Maharaj, A.~Fischer, A.~Courville,
  and Y.~Bengio.
\newblock {A Closer Look at Memorization in Deep Networks}.
\newblock In {\em Proceedings of the 34th International Conference on Machine
  Learning}, 2017.

\bibitem{geman1992neural}
S.~Geman, E.~Bienenstock, and R.~Doursat.
\newblock Neural networks and the bias/variance dilemma.
\newblock {\em Neural computation}, 4(1):1--58, 1992.

\bibitem{Neyshabur2015}
B.~Neyshabur, R.~Tomioka, and N.~Srebro.
\newblock {In search of the real inductive bias: On the role of implicit
  regularization in deep learning}.
\newblock In {\em ICLR}, 2015.

\bibitem{spigler2018jamming}
S~Spigler, M~Geiger, S~d’Ascoli, L~Sagun, G~Biroli, and M~Wyart.
\newblock A jamming transition from under-to over-parametrization affects
  generalization in deep learning.
\newblock {\em Journal of Physics A: Mathematical and Theoretical},
  52(47):474001, 2019.

\bibitem{nakkiran2020deep}
Preetum Nakkiran, Gal Kaplun, Yamini Bansal, Tristan Yang, Boaz Barak, and Ilya
  Sutskever.
\newblock Deep double descent: Where bigger models and more data hurt.
\newblock In {\em International Conference on Learning Representations}, 2020.

\bibitem{gunasekar2018characterizing}
S.~Gunasekar, J.~Lee, D.~Soudry, and N.~Srebro.
\newblock Characterizing implicit bias in terms of optimization geometry.
\newblock In {\em International Conference on Machine Learning}, pages
  1832--1841. PMLR, 2018.

\bibitem{gunasekar2018implicit}
S.~Gunasekar, J.~D. Lee, D.~Soudry, and N.~Srebro.
\newblock Implicit bias of gradient descent on linear convolutional networks.
\newblock In S.~Bengio, H.~Wallach, H.~Larochelle, K.~Grauman, N.~Cesa-Bianchi,
  and R.~Garnett, editors, {\em Advances in Neural Information Processing
  Systems}, volume~31. Curran Associates, Inc., 2018.

\bibitem{Soudry2018}
D.~Soudry, E.~Hoffer, and N.~Srebro.
\newblock The implicit bias of gradient descent on separable data.
\newblock In {\em International Conference on Learning Representations}, 2018.

\bibitem{ji2019implicit}
Z.~Ji and M.~Telgarsky.
\newblock The implicit bias of gradient descent on nonseparable data.
\newblock In {\em Conference on Learning Theory}, pages 1772--1798. PMLR, 2019.

\bibitem{arora2019implicit}
S.~Arora, N.~Cohen, W.~Hu, and Y.~Luo.
\newblock Implicit regularization in deep matrix factorization.
\newblock In H.~Wallach, H.~Larochelle, A.~Beygelzimer, F.~d\textquotesingle
  Alch\'{e}-Buc, E.~Fox, and R.~Garnett, editors, {\em Advances in Neural
  Information Processing Systems}, volume~32. Curran Associates, Inc., 2019.

\bibitem{chizat2020implicit}
L.~Chizat and F.~Bach.
\newblock Implicit bias of gradient descent for wide two-layer neural networks
  trained with the logistic loss.
\newblock In {\em Conference on Learning Theory}, pages 1305--1338. PMLR, 2020.

\bibitem{advani2020highdimensional}
M.S. Advani, A.M. Saxe, and H.~Sompolinsky.
\newblock High-dimensional dynamics of generalization error in neural networks.
\newblock {\em Neural Networks}, 132:428 -- 446, 2020.

\bibitem{neal2018modern}
B.~Neal, S.~Mittal, A.~Baratin, V.~Tantia, M.~Scicluna, S.~Lacoste-Julien, and
  I.~Mitliagkas.
\newblock A modern take on the bias-variance tradeoff in neural networks.
\newblock {\em arXiv preprint arXiv:1810.08591}, 2018.

\bibitem{mei2019generalization}
S.~Mei and A.~Montanari.
\newblock The generalization error of random features regression: Precise
  asymptotics and the double descent curve.
\newblock {\em Communications on Pure and Applied Mathematics}, 2019.

\bibitem{belkin2019reconciling}
M.~Belkin, D.~Hsu, S.~Ma, and S.~Mandal.
\newblock Reconciling modern machine-learning practice and the classical
  bias--variance trade-off.
\newblock {\em Proceedings of the National Academy of Sciences},
  116(32):15849--15854, 2019.

\bibitem{hastie2022surprises}
T.~Hastie, A.~Montanari, S.~Rosset, and R.J. Tibshirani.
\newblock Surprises in high-dimensional ridgeless least squares interpolation.
\newblock {\em The Annals of Statistics}, 50(2):949--986, 2022.

\bibitem{dascoli2020double}
S.~d'Ascoli, M.~Refinetti, G.~Biroli, and F.~Krzakala.
\newblock Double trouble in double descent : Bias and variance(s) in the lazy
  regime.
\newblock In {\em ICML}, 2020.

\bibitem{adlam2020understanding}
B.~Adlam and J.~Pennington.
\newblock Understanding double descent requires a fine-grained bias-variance
  decomposition.
\newblock In H.~Larochelle, M.~Ranzato, R.~Hadsell, M.~F. Balcan, and H.~Lin,
  editors, {\em Advances in Neural Information Processing Systems}, volume~33,
  pages 11022--11032. Curran Associates, Inc., 2020.

\bibitem{lin2021causes}
L.~Lin and E.~Dobriban.
\newblock What causes the test error? going beyond bias-variance via anova.
\newblock {\em J. Mach. Learn. Res.}, 22:155--1, 2021.

\bibitem{geiger2020scaling}
M.~Geiger, A.~Jacot, S.~Spigler, F.~Gabriel, L.~Sagun, S.~d’Ascoli,
  G.~Biroli, C.~Hongler, and M.~Wyart.
\newblock Scaling description of generalization with number of parameters in
  deep learning.
\newblock {\em Journal of Statistical Mechanics: Theory and Experiment},
  2020(2):023401, 2020.

\bibitem{bartlett2020benign}
P.L. Bartlett, P.M. Long, G.~Lugosi, and A.~Tsigler.
\newblock Benign overfitting in linear regression.
\newblock {\em Proceedings of the National Academy of Sciences},
  117(48):30063--30070, 2020.

\bibitem{liu2020bad}
S.~Liu, D.~Papailiopoulos, and D.~Achlioptas.
\newblock Bad global minima exist and sgd can reach them.
\newblock In H.~Larochelle, M.~Ranzato, R.~Hadsell, M.~F. Balcan, and H.~Lin,
  editors, {\em Advances in Neural Information Processing Systems}, volume~33,
  pages 8543--8552. Curran Associates, Inc., 2020.

\bibitem{huang2017densely}
G.~Huang, Z.~Liu, L.~Van Der~Maaten, and K.~Q. Weinberger.
\newblock Densely connected convolutional networks.
\newblock In {\em Proceedings of the IEEE conference on computer vision and
  pattern recognition}, pages 4700--4708, 2017.

\bibitem{krizhevsky2009learning}
A.~Krizhevsky, G.~Hinton, et~al.
\newblock Learning multiple layers of features from tiny images.
\newblock https://www.cs.toronto.edu/~kriz/learning-features-2009-TR.pdf, 2009.

\bibitem{zagoruyko2016wide}
S.~Zagoruyko and N.~Komodakis.
\newblock Wide residual networks.
\newblock {\em arXiv preprint arXiv:1605.07146}, 2016.

\bibitem{imagenet_cvpr09}
J.~Deng, W.~Dong, R.~Socher, L.-J. Li, K.~Li, and L.~Fei-Fei.
\newblock {ImageNet: A Large-Scale Hierarchical Image Database}.
\newblock In {\em CVPR09}, 2009.

\bibitem{lecun1998}
Y.~LeCun and C.~Cortes.
\newblock {The MNIST database of handwritten digits}, 1998.

\bibitem{he2016deep}
K.~He, X.~Zhang, S.~Ren, and J.~Sun.
\newblock Deep residual learning for image recognition.
\newblock In {\em Proceedings of the IEEE conference on computer vision and
  pattern recognition}, pages 770--778, 2016.

\bibitem{hastie01statisticallearning}
T.~Hastie, R.~Tibshirani, and J.~Friedman.
\newblock {\em The Elements of Statistical Learning}.
\newblock Springer Series in Statistics. Springer New York Inc., New York, NY,
  USA, 2001.

\bibitem{bengio2013representation}
Y.~Bengio, A.~Courville, and P.~Vincent.
\newblock Representation learning: A review and new perspectives.
\newblock {\em IEEE transactions on pattern analysis and machine intelligence},
  35(8):1798--1828, 2013.

\bibitem{ansuini2019intrinsic}
A.~Ansuini, A.~Laio, J.~H. Macke, and D.~Zoccolan.
\newblock Intrinsic dimension of data representations in deep neural networks.
\newblock In {\em Advances in Neural Information Processing Systems}, pages
  6109--6119, 2019.

\bibitem{hestness2017deep}
J.~Hestness, S.~Narang, N.~Ardalani, G.~Diamos, H.~Jun, H.~Kianinejad,
  M.~Patwary, M.~Ali, Y.~Yang, and Y.~Zhou.
\newblock Deep learning scaling is predictable, empirically.
\newblock {\em arXiv:1712.00409}, 2017.

\bibitem{rosenfeld2020constructive}
J.S. Rosenfeld, A.~Rosenfeld, Y.~Belinkov, and N.~Shavit.
\newblock A constructive prediction of the generalization error across scales.
\newblock In {\em International Conference on Learning Representations}, 2020.

\bibitem{kaplan2020scaling}
J.~Kaplan, S.~McCandlish, T.~Henighan, T.B. Brown, B.~Chess, R.~Child, S.~Gray,
  A.~Radford, J.~Wu, and D.~Amodei.
\newblock Scaling laws for neural language models.
\newblock {\em arXiv preprint arXiv:2001.08361}, 2020.

\bibitem{henighan2020scaling}
T.~Henighan, J.~Kaplan, M.~Katz, M.~Chen, C.~Hesse, J.~Jackson, H.~Jun, T.B.
  Brown, P.~Dhariwal, S.~Gray, et~al.
\newblock Scaling laws for autoregressive generative modeling.
\newblock {\em arXiv preprint arXiv:2010.14701}, 2020.

\bibitem{bahri2021explaining}
Y.~Bahri, E.~Dyer, J.~Kaplan, J.~Lee, and U.~Sharma.
\newblock Explaining neural scaling laws.
\newblock {\em arXiv preprint arXiv:2102.06701}, 2021.

\bibitem{jacot2018neural}
A.~Jacot, F.~Gabriel, and C.~Hongler.
\newblock Neural tangent kernel: Convergence and generalization in neural
  networks.
\newblock In {\em Advances in Neural Information Processing Systems 32}, pages
  8571--8580, 2018.

\bibitem{du2018gradient}
S.~Du, J.~Lee, Y.~Tian, A.~Singh, and B.~Poczos.
\newblock Gradient descent learns one-hidden-layer {CNN}: Don’t be afraid of
  spurious local minima.
\newblock In {\em Proceedings of the 35th International Conference on Machine
  Learning}, volume~80, pages 1339--1348, 2018.

\bibitem{chizat2019lazy}
L.~Chizat, E.~Oyallon, and F.~Bach.
\newblock On lazy training in differentiable programming.
\newblock In {\em Advances in Neural Information Processing Systems}, pages
  2937--2947, 2019.

\bibitem{geiger2020disentangling}
M.~Geiger, S.~Spigler, A.~Jacot, and M.~Wyart.
\newblock Disentangling feature and lazy training in deep neural networks.
\newblock {\em Journal of Statistical Mechanics: Theory and Experiment},
  2020(11):113301, 2020.

\bibitem{Mei2018}
S.~Mei, A.~Montanari, and P.~Nguyen.
\newblock {A mean field view of the landscape of two-layer neural networks}.
\newblock {\em Proceedings of the National Academy of Sciences},
  115(33):E7665--E7671, 2018.

\bibitem{rotskoff2018interacting}
G.M. Rotskoff and E.~Vanden-Eijnden.
\newblock {Parameters as interacting particles: long time convergence and
  asymptotic error scaling of neural networks}.
\newblock In {\em Advances in Neural Information Processing Systems 31}, pages
  7146--7155, 2018.

\bibitem{Chizat2018}
L.~Chizat and F.~Bach.
\newblock On the global convergence of gradient descent for over-parameterized
  models using optimal transport.
\newblock In {\em Advances in Neural Information Processing Systems 31}, pages
  3040--3050, 2018.

\bibitem{Sirignano2018}
J.~Sirignano and K.~Spiliopoulos.
\newblock {Mean field analysis of neural networks: A central limit theorem}.
\newblock {\em Stochastic Processes and their Applications}, 2019.

\bibitem{goldt2019dynamics}
S.~Goldt, M.S. Advani, A.M. Saxe, F.~Krzakala, and L.~Zdeborov{\'a}.
\newblock Dynamics of stochastic gradient descent for two-layer neural networks
  in the teacher-student setup.
\newblock In {\em Advances in Neural Information Processing Systems 32}, 2019.

\bibitem{refinetti2021classifying}
M.~Refinetti, S.~Goldt, F.~Krzakala, and L.~Zdeborova.
\newblock Classifying high-dimensional gaussian mixtures: Where kernel methods
  fail and neural networks succeed.
\newblock In Marina Meila and Tong Zhang, editors, {\em Proceedings of the 38th
  International Conference on Machine Learning}, volume 139 of {\em Proceedings
  of Machine Learning Research}, pages 8936--8947. PMLR, 18--24 Jul 2021.

\bibitem{kuditipudi2019explaining}
R.~Kuditipudi, X.~Wang, H.~Lee, Y.~Zhang, Z.~Li, W.~Hu, R.~Ge, and S.~Arora.
\newblock Explaining landscape connectivity of low-cost solutions for
  multilayer nets.
\newblock In H.~Wallach, H.~Larochelle, A.~Beygelzimer, F.~d\textquotesingle
  Alch\'{e}-Buc, E.~Fox, and R.~Garnett, editors, {\em Advances in Neural
  Information Processing Systems}, volume~32. Curran Associates, Inc., 2019.

\bibitem{shevchenko2020landscape}
A.~Shevchenko and M.~Mondelli.
\newblock Landscape connectivity and dropout stability of sgd solutions for
  over-parameterized neural networks.
\newblock In {\em International Conference on Machine Learning}, pages
  8773--8784. PMLR, 2020.

\bibitem{nguyen2021connectivity}
Q.~Nguyen, P.~Brechet, and M.~Mondelli.
\newblock On connectivity of solutions in deep learning: The role of
  over-parameterization and feature quality.
\newblock In {\em Advances in Neural Information Processing Systems},
  volume~34. Curran Associates, Inc., 2021.

\bibitem{hinton2012improving}
G.~E. Hinton, N.~Srivastava, A.~Krizhevsky, I.~Sutskever, and R.~Salakhutdinov.
\newblock Improving neural networks by preventing co-adaptation of feature
  detectors.
\newblock {\em arXiv:1207.0580}, 2012.

\bibitem{srivastava2014dropout}
N.~Srivastava, G.~Hinton, A.~Krizhevsky, I.~Sutskever, and R.~Salakhutdinov.
\newblock Dropout: a simple way to prevent neural networks from overfitting.
\newblock {\em The journal of machine learning research}, 15(1):1929--1958,
  2014.

\bibitem{micikevicius2018mixed}
P.~Micikevicius, S.~Narang, J.~Alben, G.~Diamos, E.~Elsen, D.~Garcia,
  B.~Ginsburg, M.~Houston, O.~Kuchaiev, G.~Venkatesh, and H.~Wu.
\newblock Mixed precision training.
\newblock In {\em International Conference on Learning Representations}, 2018.

\bibitem{bello2021revisiting}
I.~Bello, W.~Fedus, X.~Du, E.~D. Cubuk, A.~Srinivas, T.-Y. Lin, J.~Shlens, and
  B.~Zoph.
\newblock Revisiting resnets: Improved training and scaling strategies.
\newblock In M.~Ranzato, A.~Beygelzimer, Y.~Dauphin, P.S. Liang, and J.~Wortman
  Vaughan, editors, {\em Advances in Neural Information Processing Systems},
  volume~34, pages 22614--22627. Curran Associates, Inc., 2021.

\bibitem{goyal2017accurate}
P.~Goyal, P.~Doll{\'a}r, R.~Girshick, P.~Noordhuis, L.~Wesolowski, A.~Kyrola,
  A.~Tulloch, Y.~Jia, and K.~He.
\newblock Accurate, large minibatch sgd: Training imagenet in 1 hour.
\newblock {\em arXiv preprint arXiv:1706.02677}, 2017.

\bibitem{yang2020rethinking}
Zitong Yang, Yaodong Yu, Chong You, Jacob Steinhardt, and Yi~Ma.
\newblock Rethinking bias-variance trade-off for generalization of neural
  networks.
\newblock In Hal~Daumé III and Aarti Singh, editors, {\em Proceedings of the
  37th International Conference on Machine Learning}, volume 119 of {\em
  Proceedings of Machine Learning Research}, pages 10767--10777. PMLR, 13--18
  Jul 2020.

\end{thebibliography}

\section{NeurIPS checklist}

\begin{enumerate}

\item For all authors...
\begin{enumerate}
  \item Do the main claims made in the abstract and introduction accurately reflect the paper's contributions and scope?
    \answerYes{}
  \item Did you describe the limitations of your work?
    \answerYes{}
  \item Did you discuss any potential negative societal impacts of your work?
    \answerYes{}
  \item Have you read the ethics review guidelines and ensured that your paper conforms to them?
    \answerYes{}
\end{enumerate}

\item If you are including theoretical results...
\begin{enumerate}
  \item Did you state the full set of assumptions of all theoretical results?
    \answerNA{}
        \item Did you include complete proofs of all theoretical results?
    \answerNA{}
\end{enumerate}

\item If you ran experiments...
\begin{enumerate}
  \item Did you include the code, data, and instructions needed to reproduce the main experimental results (either in the supplemental material or as a URL)?
    \answerYes{We provide code to reproduce all our experiments and figures at ~\url{https://anonymous.4open.science/r/redundant_representation2022-E4FD}.}
  \item Did you specify all the training details (e.g., data splits, hyperparameters, how they were chosen)?
    \answerYes{See Methods section.}
        \item Did you report error bars (e.g., with respect to the random seed after running experiments multiple times)?
    \answerYes{}
        \item Did you include the total amount of compute and the type of resources used (e.g., type of GPUs, internal cluster, or cloud provider)?
    \answerYes{We ran all our experiments on V100 GPUs}
\end{enumerate}

\item If you are using existing assets (e.g., code, data, models) or curating/releasing new assets...
\begin{enumerate}
  \item If your work uses existing assets, did you cite the creators?
    \answerYes{We used the standard MNIST~\cite{lecun1998}, CIFAR10-CIFAR100 data sets~\cite{krizhevsky2009learning} and ImageNet~\cite{imagenet_cvpr09}. See Methods section}
  \item Did you mention the license of the assets?
    \answerNA{}
  \item Did you include any new assets either in the supplemental material or as a URL?
    \answerNA{}
  \item Did you discuss whether and how consent was obtained from people whose data you're using/curating?
    \answerNA{}
  \item Did you discuss whether the data you are using/curating contains personally identifiable information or offensive content?
    \answerNA{}
\end{enumerate}

\item If you used crowdsourcing or conducted research with human subjects...
\begin{enumerate}
  \item Did you include the full text of instructions given to participants and screenshots, if applicable?
    \answerNA{}
  \item Did you describe any potential participant risks, with links to Institutional Review Board (IRB) approvals, if applicable?
    \answerNA{}
  \item Did you include the estimated hourly wage paid to participants and the total amount spent on participant compensation?
    \answerNA{}
\end{enumerate}

\end{enumerate}

\newpage
\appendix
\onecolumn
\renewcommand{\thefigure}{S\arabic{figure}}
\setcounter{figure}{0}
\section{Hyperparameters used and training procedures}
\label{sec:training_hyperparams}

\paragraph{Fully-connected networks on MNIST.}
We train the fully-connected networks for 5000 epochs with stochastic gradient descent using the following hyperparameters: batch size = 256, momentum = 0.9, learning rate = $10^{-3}$, weight decay = $10^{-2}$.
We optimize our networks using Adam.

\paragraph{Wide-ResNet-28 and DenseNet40-BC on CIFAR10/100.} All the models are trained for 200 epochs with stochastic gradient descent with a batch size = 128, momentum = 0.9, and cosine annealing scheduler starting with a learning rate of 0.1. The training set is augmented with horizontal flips with 50\% probability and random cropping the images padded with four pixels on each side.  
On CIFAR10 trained on WR28 we select a weight decay equal to $5 \cdot 10^{-4}$ and label smoothing magnitudes equal to 0.1 for WR28-\{0.25, 0.5, 1, 2\} and equal to 0 for WR28-\{4, 8\}. 
On CIFAR10 trained on Densenet40-BC we set a weight decay equal to $5 \cdot 10^{-4}$ and label smoothing magnitudes equal to 0.05 for all the networks 
On CIFAR100 trained on WR28 we set weight decays equal to \{10, 7, 5, 5, 5\}$\cdot 10^{-4}$ and label smoothing magnitudes equal to \{0.1, 0.07, 0.05, 0, 0\} for WR28-\{1, 2, 4, 8, 16\} respectively.
All the hyperparameters were selected with a small  grid search.

\paragraph{ResNet50 on ImageNet.}
We train all the ResNet50 with mixed precision
\citep{micikevicius2018mixed} for 120 epochs with a weight decay of
$4\cdot10^{-5}$ and label smoothing rate of 0.1 \citep{bello2021revisiting}.
The input size is $224\times 224$ and the training set is augmented with random crops and horizontal flips with 50\% probability. The per-GPU batch size is set to 128 and is halved for the widest networks to fit in the GPU memory. 
The networks are trained on 8 or 16 Volta V100 GPUs so as to keep the batch size $B$ equal to 1024. The learning rate is increased linearly from 0 to 0.1$\cdot B$/256 \citep{goyal2017accurate} for the first five epochs and then annealed to zero with a cosine schedule.

\begin{table}[h!]
  \caption{Test accuracy (average over four runs)}
  \label{table accuracy}
  \centering
\begin{center}
  \begin{tabular}{llllll}
    \toprule
    \multicolumn{2}{c}{CIFAR10} &     \multicolumn{2}{c}{CIFAR100} &  \multicolumn{2}{c}{ImageNet (top1)}  \\
    \cmidrule(r){1-6} 
    network     & accuracy & network  & accuracy & network  &  accuracy\\
    \midrule
    Wide-RN28-0.25 & 84.1 &    Wide-RN28-1  & 70.4  &     RN50-0.25 & 67.0   \\
    Wide-RN28-0.5 & 90.3 &     Wide-RN28-2  & 75.7  &     RN50-0.5  & 74.1 \\
    Wide-RN28-1 & 93.4 &       Wide-RN28-4  & 79.6  &     RN50-1  & 77.6 \\    
    Wide-RN28-2 & 95.2 &       Wide-RN28-8  & 80.8  &     RN50-2  & 79.1 \\ 
    Wide-RN28-4  & 95.9  &     Wide-RN28-16 & 81.9 &      RN50-4  & 79.5 \\ 
    Wide-RN28-8  & 96.1  \\
    DenseNet40-BC (k=8)  & 91.6  \\
    DenseNet40-BC (k=16)   & 93.9  \\
    DenseNet40-BC (k=32)   & 95.1 \\
    DenseNet40-BC (k=64)   & 95.7 \\
    DenseNet40-BC (k=128)   & 96.0 \\ 
    
    \bottomrule
  \end{tabular}
  \end{center}
  
\end{table} 
\newpage
\section{Additional experiments}
\label{sec:additional_experiments}
\begin{figure}[h!]
\centering
\includegraphics[width=0.9\columnwidth]{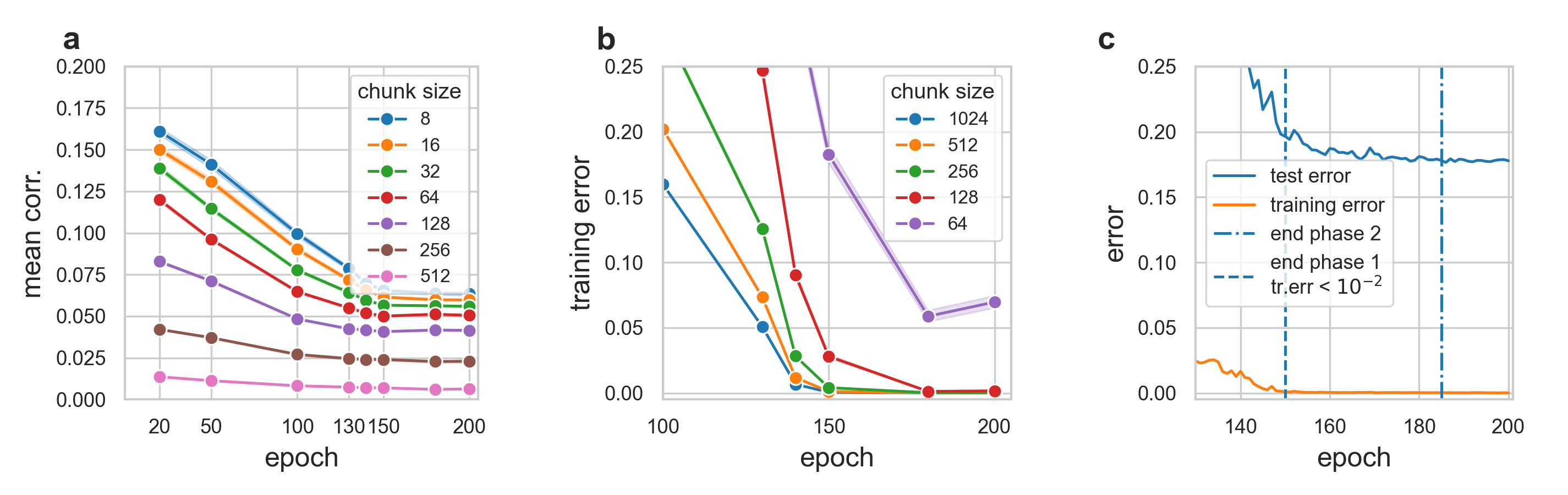}
\caption{\label{fig:app_cifar100_dyn}
\textbf{Training dynamics on CIFAR100.}
\textbf{a:}As in Fig.~\ref{fig:hallmarks-redundancy}, we show the mean correlation of the residuals of the linear reconstruction of the final representation of a Wide-Resnet28-8 from chunks, but this time as a function of training epochs. A small correlation indicates that the reconstruction error in going from chunks to the final representation can be modeled as independent noise.
\textbf{b:} Training error of chunks of a Wide-Resnet28-8 and its full layer representation. 
From epoch 150 to epoch 185 the training error of the chunks with size 128/256 decreases below 0.5\%, while for smaller chunk sizes it remains above 5\%. Random chunks with sizes larger than 128/256 can fit the training set, thus having the same representational power as the whole network on the training data.
For W > 128/256 the test accuracy is decaying approximately with the same law as that of independent networks with the same width (see Fig. \ref{fig:chunk-error-scaling}). This picture suggests that for CIFAR100 the size of a clone is 128/256, slightly larger than the size of the clones in CIFAR10.
\textbf{c:} Training and test error dynamics for the same Wide-ResNet28-8.
After epoch 150 the training error of the full network remains consistently smaller than 0.1\% (orange profile) while the test error continues to decrease until epoch 185 from 0.194 to 0.1765 (blue profile).  
In the same range of epochs (150-185) the training error of smaller chunks decreases sensibly (see panel \textbf{b}). 
}
\end{figure}

%
%
%

%
%
%
%
\begin{figure}[!b]
\centering
\includegraphics[width=0.9\columnwidth]{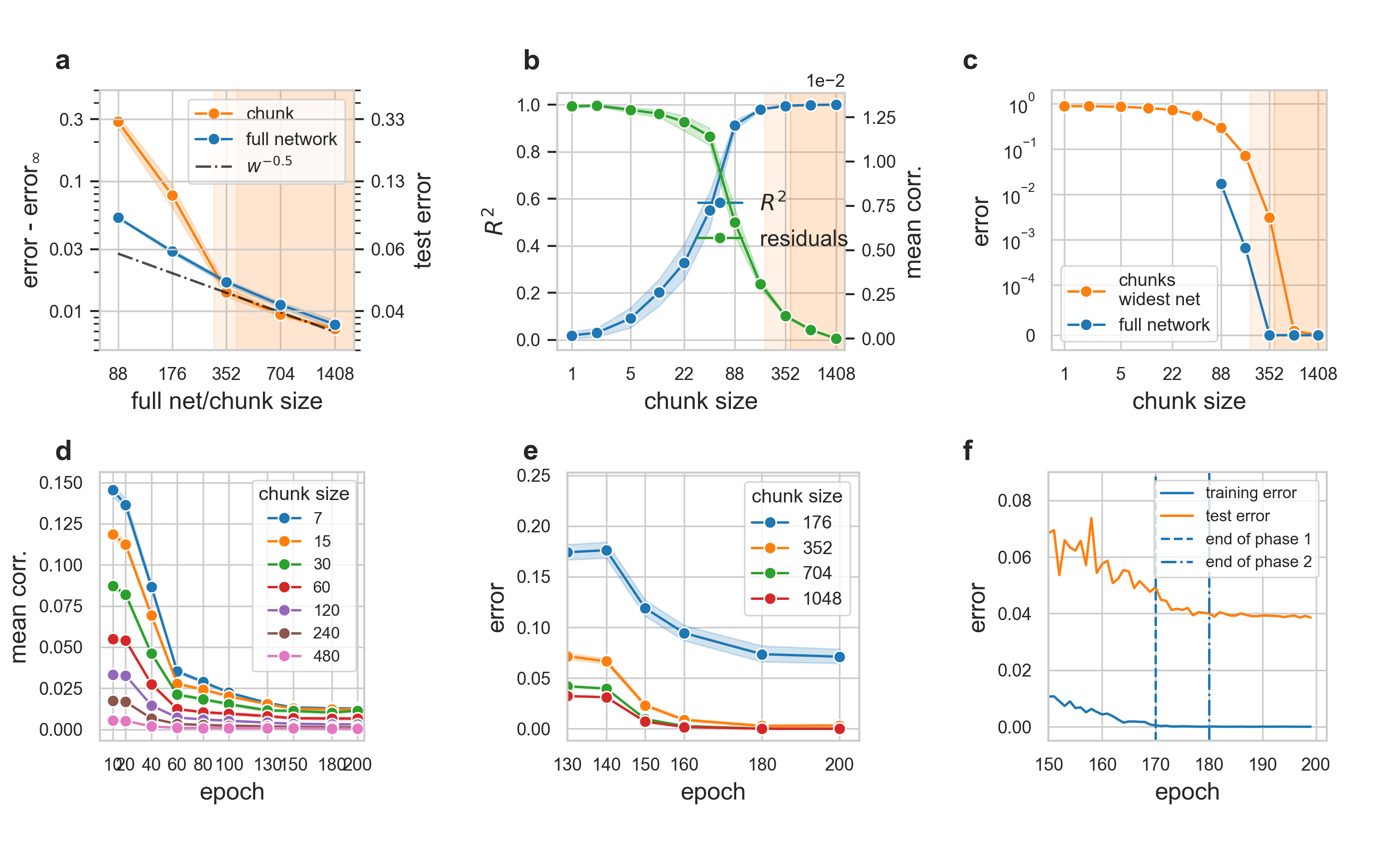}
\caption{\label{fig:app_densenet} 
\textbf{A DenseNet40 architecture.}
\textbf{a:} Decay of the test error of independent networks (blue) and chunks of the widest network (orange) to the error of an ensemble average of ten of the widest networks (DenseNet40-BC, k=128)
\textbf{b:} 
Blue profile: $R^2$ coefficient of the ridge regression of a chunk of $w_c$ neurons ($x$-axis) to the full layer representation.
Green profile: mean correlation of the residuals of the mapping as described in Sec. \ref{sec:methods-analysis}.
\textbf{c:} 
Training error of various DenseNet40 of increasing width (blue) and of chunks of the widest architecture (orange).
\textbf{d:} 
The mean correlation of the residuals from the linear reconstruction of the final representation from chunks of a given size for a DenseNet40-BC (k=128) during training. 
\textbf{e:} 
Training error dynamics of chunks of a DenseNet40-BC (k=128).
\textbf{f:} 
Training and test error dynamics for a DenseNet40-BC (k=128).
}
\end{figure}
\begin{figure}[!b]
\centering
\includegraphics[width=0.8\columnwidth]{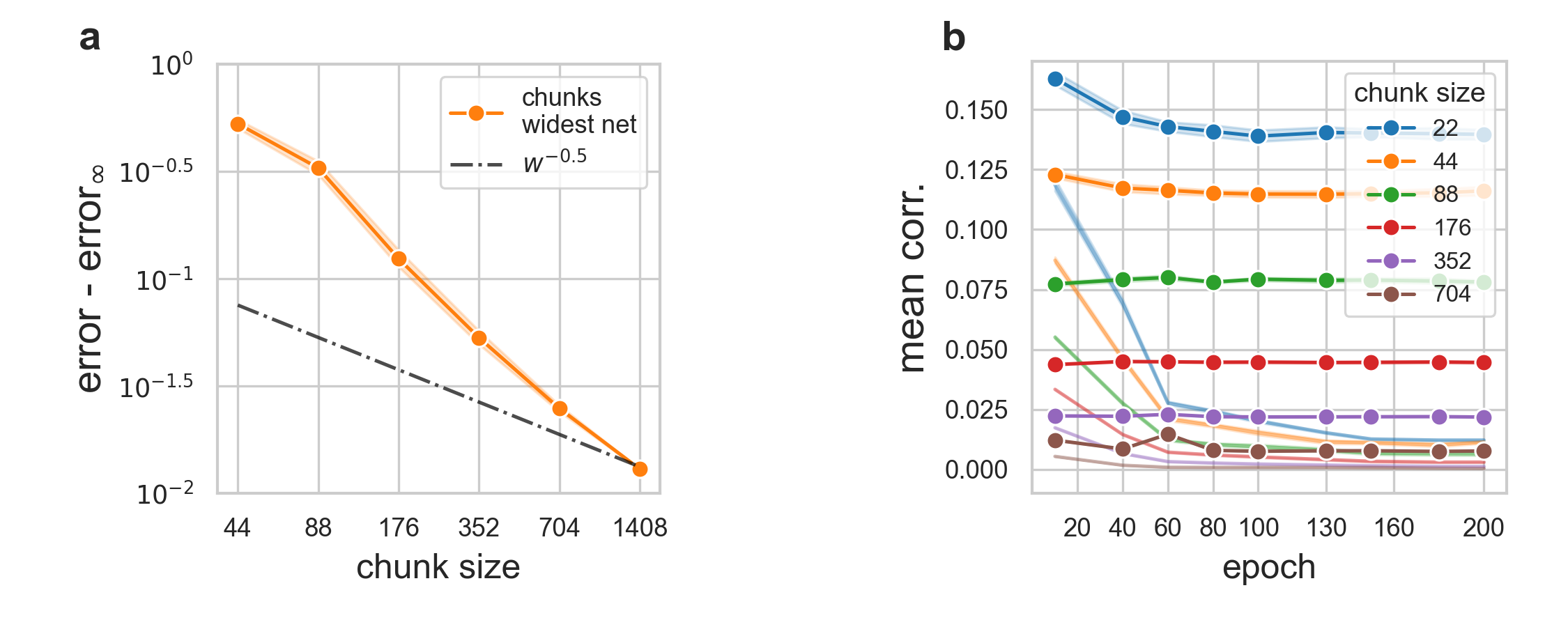}
\caption{\label{fig:app_densenet_not_reg} \textbf{A Densenet40 not regularized.} A DenseNet40-BC (k=128) trained on CIFAR10 without weight decay. This experiment reproduces on a DenseNet the analysis shown on a Wide-ResNet28 in Sec. \ref{sec:results}. It shows that \textbf{a}: also in a DenseNet architecture not well regularized $\mathrm{error}$ -$\mathrm{error}_\infty$ decays faster than $w_c^{\nicefrac{-1}{2}}$  and \textbf{b}: the mean correlation of the residuals do not decrease during training. The thin profiles of panel \textbf{b} are the same as those shown in Fig. \ref{fig:app_densenet}-d.}
\end{figure}

\label{ref:app_ony_last}
\begin{figure}[!b]
\centering
\includegraphics[width=0.8\columnwidth]{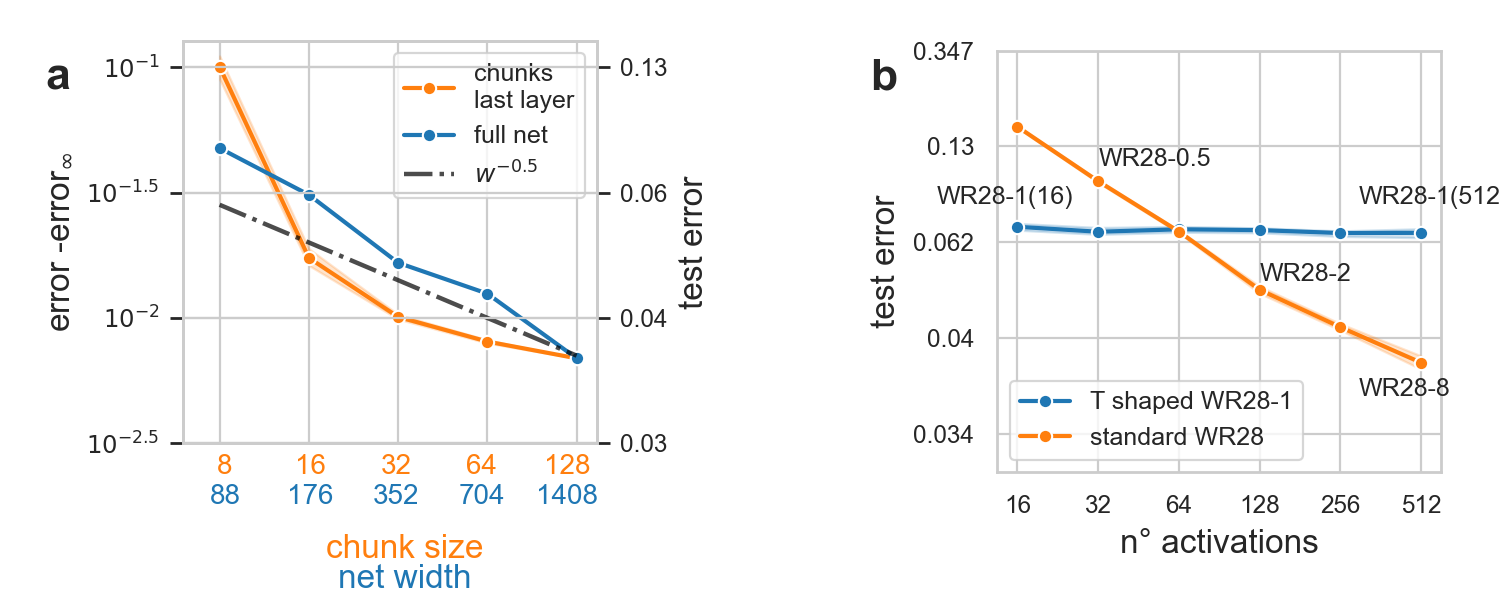}
\caption{\label{fig:app_only_last} 
\textbf{Impact of the width of the intermediate layers.}
We study how the scaling of the test error is affected (Fig. {\bfseries a}) by increasing the width of the intermediate representations while keeping the width of the last layer constant or (Fig. {\bfseries b}) by  increasing the last layer width while keeping the width of the network constant.
In \ref{fig:app_only_last}-{\bfseries a} we trained DenseNet40 on CIFAR10 with an additional 1 × 1 convolution to keep the number of output channels fixed at 128. Figure \ref{fig:app_only_last}-{\bfseries a } shows that increasing the width of intermediate layers makes the test accuracy of the full network decay approximately as $w_c^{-\nicefrac{1}{2}}$, even when the width of the final representation is fixed.
A bottleneck of 128 channels makes the clones much smaller: 
The orange profile shows that a strong deviation from the $w^{\nicefrac{-1}{2}}$ can be seen for chunk sizes smaller than 16 (vs 350 Fig. 1b, main paper). We also verified that 16 random neurons are sufficient to interpolate the training set (error $<5 \cdot 10^{-3}$) and that the $R^2$ coefficient of fit to the full layer is 0.912 (0.98 for chunk sizes $=32$). The phenomenology described in the manuscript applies also when a bottleneck of 128 channels is added at the end of the network.
In a second experiment, we trained a ResNet28-1 increasing only the number of channels in the last layer. 
We modified the number of output channels of the last block of conv4 and analyzed the representation after average pooling, as we did in the other experiments. 
The network was trained for 200 epochs using the same hyperparameters and protocol described in Sec. \ref{sec:methods}. 
Figure \ref{fig:app_only_last}-{\bfseries b} shows that the test error of the modified ResNet28-1 is approximately constant (blue profile). On the contrary, when we increase the width of the whole network the test error decays to the asymptotic test error with an approximate scaling of  $1/\sqrt{w}$ (orange profile).}
\end{figure}

\begin{figure}[!b]
\centering
\includegraphics[width=1.\columnwidth]{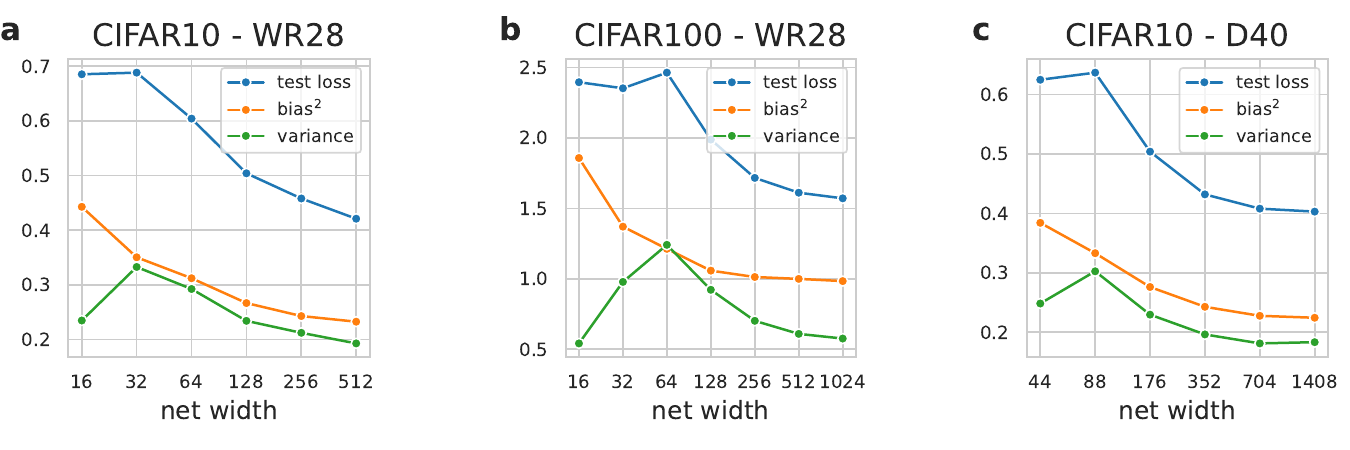}
\caption{\label{fig:app_bias_variance_ls0} \textbf{Bias-variance profiles in CIFRA10 and CIFAR100.} 
We compute the bias and the variance profiles for the convolutional architectures analyzed in the paper: Wide-ResNets and DenseNets trained on CIFAR10,  and Wide-ResNets trained on CIFAR100. Since we trained the models using the cross-entropy loss, the standard bias-variance decomposition, which assumes the square loss, does not apply. Instead, we used the method recently proposed by Yang \emph{et al.}~\cite{yang2020rethinking} to estimate the bias and the variance on networks trained with cross-entropy loss. 
The average over the data distribution is approximated by splitting the CIFAR training sets into five disjoint subsets containing 10~000 images each and training the networks from scratch on each of them. We use the same regularization for all the networks, namely that of the largest architectures, with weight decay equal to $5 \cdot 10^{-4}$ and label smoothing equal to 0.
We repeat the procedure 4 times, for a total of 20 training runs for each network width, as described in Ref.~\cite{yang2020rethinking}.
We show the test loss curves as well as the squared bias and variance. As expected, the bias of the models decreases as we add parameters and make the model more flexible. 
The variance of the models initially grows with width to reach its peak at $W_\mathrm{peak} = 32$ and $64$ for CIFAR10 and CIFAR100 trained on Wide-ResNet28 (a, b) and  $W_\mathrm{peak} = 88$ on CIFAR10 trained on DenseNet40 (c). 
As we increase the width, the variance decreases, allowing the model to generalize better and better and defying the classical bias-variance trade-off. 
The clone size $w_c^*$ for these architectures are slightly above the widths at which the variance peaks and are $w_c^* = 64$, and $128$ for CIFAR10 and CIFAR100 trained on Wide-ResNet28 (compare Figs.~\ref{fig:chunk-error-scaling} and~\ref{fig:hallmarks-redundancy}) and 
$w_c^* = 170/250$ (Fig.~\ref{fig:app_densenet}). 
In all cases, the onset of the clones occurs at a width that is approximately two times larger than $W_\mathrm{peak}$, similar to the width at which an architecture of size $w_c^*$ interpolates the training set.
%
%
%
%
%
}
\end{figure}

\end{document}